\documentclass[10pt,twocolumn,letterpaper]{article}

\usepackage{iccv}
\usepackage{times}
\usepackage{epsfig}
\usepackage{graphicx}
\usepackage{amsmath}
\usepackage{amssymb}
\usepackage{multirow}
\usepackage[inline]{enumitem}
\usepackage{dsfont}
\usepackage[dvipsnames]{xcolor}
\usepackage{multirow}

\usepackage[pagebackref=true,breaklinks=true,colorlinks,bookmarks=false]{hyperref}

\usepackage{pifont}
\newcommand{\cmark}{\ding{51}}%
\iccvfinalcopy 

\def\iccvPaperID{2209} 

\begin{document}

\title{Delving Deep Into Hybrid Annotations for 3D Human Recovery in the Wild}

\author{
	Yu Rong$^{1}$ \hspace{9pt} Ziwei Liu$^{1}$ \hspace{9pt} Cheng Li$^{2}$ \hspace{9pt} Kaidi Cao$^{4}$ \hspace{8pt}  Chen Change Loy$^{3}$\\
	\small{$^{1}$CUHK - SenseTime Joint Lab, The Chinese University of Hong Kong}\\
	\small{$^{2}$SenseTime Research \hspace{9pt} $^{3}$Nanyang Technological University \hspace{9pt} $^{4}$Stanford University}\\
	{\tt\small \{ry017, zwliu\}@ie.cuhk.edu.hk  \hspace{3pt} chengli@sensetime.com} \\
	{\tt\small kaidicao@cs.stanford.edu \hspace{3pt} ccloy@ntu.edu.sg}	
}

\maketitle


\begin{abstract}	
Though much progress has been achieved in single-image 3D human recovery, estimating 3D model for in-the-wild images remains a formidable challenge. The reason lies in the fact that obtaining high-quality 3D annotations for in-the-wild images is an extremely hard task that consumes enormous amount of resources and manpower.  
To tackle this problem, previous methods adopt a hybrid training strategy that exploits multiple heterogeneous types of annotations including 3D and 2D while leaving the efficacy of each annotation not thoroughly investigated.
In this work, we aim to perform a comprehensive study on cost and effectiveness trade-off between different annotations.
Specifically, we focus on the challenging task of in-the-wild 3D human recovery from single images when paired 3D annotations are not fully available.
Through extensive experiments, we obtain several observations: 1) 3D annotations are efficient, whereas traditional 2D annotations such as 2D keypoints and body part segmentation are less competent in guiding 3D human recovery. 2) Dense Correspondence such as DensePose~\cite{densepose} is effective. When there are no paired in-the-wild 3D annotations available, the model exploiting dense correspondence can achieve 92\% of the performance compared to a model trained with paired 3D data.
We show that incorporating dense correspondence into in-the-wild 3D human recovery is promising and competitive due to its high efficiency and relatively low annotating cost. Our model trained with dense correspondence can serve as a strong reference for future research
\footnote{Code and models are available at the project page: \url{https://penincillin.github.io/dct_iccv2019}}.

\end{abstract}


\section{Introduction}

Recovering 3D human model~\cite{sun2018im2avatar,kanazawa2018end} is essential in many applications such as augmented reality.
Recent studies~\cite{tan2017indirect,kanazawa2018end,omran2018neural,pavlakos2018learning} typically use a parametric model known as Skinned Multi-Person Linear Model (SMPL)~\cite{loper2015smpl} to represent 3D human models and estimate parameters of SMPL with a deep convolutional neural network (DCNN).
Training such a deep network to handle 3D human recovery in the wild is challenging, as obtaining high-quality 3D annotations in unconstrained environments for training are both laborious and expensive. To circumvent this hurdle, one often has to adopt \textit{hybrid annotations} for training, so as to leverage limited annotations from multiple datasets to avoid overfitting. For instance, Kanazawa~\etal~\cite{kanazawa2018end} train their models using both 3D joints of Human3.6M dataset~\cite{ionescu2014human3} and 2D keypoints from COCO dataset~\cite{lin2014microsoft}. Alternatively, apart from using an RGB image as an input to a network, one would introduce an auxiliary input as a prior to improve performance, \eg, Omran~\etal~\cite{omran2018neural} use body part segmentation as an intermediate representation.

\begin{figure}[t]
	\begin{center}
		\includegraphics[width=1\linewidth]{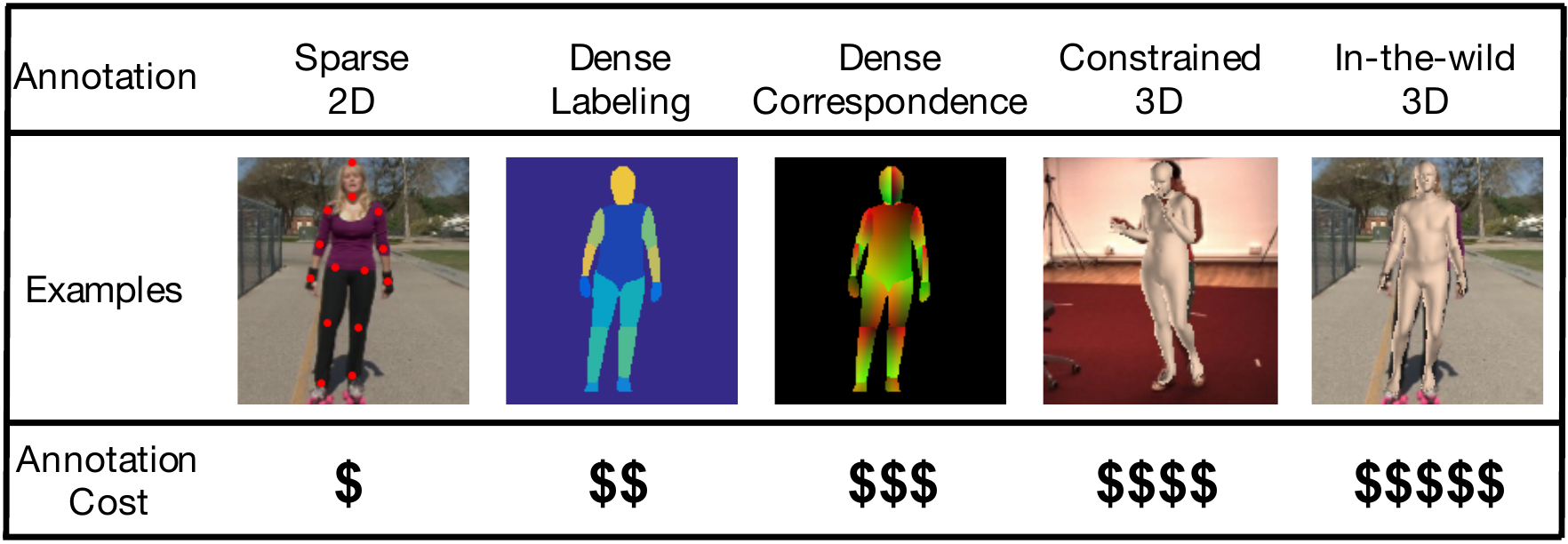}
	\end{center}
	\vspace{-0.4cm}
	\caption{\textbf{Annotations overview for 3D human recovery}. We study five kinds of annotations that are typically used in training deep networks for 3D human recovery. The number of `$\$$' indicates the annotation cost of obtaining the corresponding annotations. A higher number of `$\$$' suggests a higher cost.}
	\label{fig:anno_type}
	\vspace{-0.7cm}
\end{figure} 

As summarized in Fig.~\ref{fig:anno_type}, there are five common types of annotations:
(a) Sparse 2D annotations such as 2D keypoints,
(b) Dense labeling such as body part segmentation,
(c) Dense correspondence such as the IUV maps produced by DensePose~\cite{densepose,neverova2019slim},
(d) Constrained 3D annotations, \ie, 3D annotations for images captured in constrained environments, such as Human3.6M~\cite{ionescu2014human3}, and
(e) In-the-wild 3D annotations, \ie, 3D annotations for in-the-wild images, such as UP-3D~\cite{lassner2017unite}. 
These annotations not only vary in their expressiveness but also their labeling cost. For instance, 3D annotations like SMPL are more expressive than the dense correspondence, since the former encapsulates 3D deformable surface model while the latter only retains the UV fields. However, establishing 3D annotations requires a more complex annotation system than that required for annotating dense correspondence. Annotating dense correspondence such as DensePose~\cite{densepose} could be accomplished solely by human annotators while obtaining 3D annotations usually requires auxiliary facilities such as sparse markers~\cite{loper2014mosh} and IMUs~\cite{vonMarcard2018}.

In this study, we aim to perform a systematic study to investigate the cost and effectiveness trade-off between using different annotations in learning a deep network for 3D human recovery. We focus our study on the challenging task of recovering 3D human model from in-the-wild images, especially in the case when in-the-wild 3D annotations are insufficient, and how other annotation types could complement and bridge the gap.
Our study is conducted using a unified and simple network, which could serve as a solid baseline for future study. Two aspects of using different annotations are investigated, \ie, the effect of different annotations in serving as (a) a supervisory signal, (b) as an input to the network. 
%

Our experiments reveal several observations:

\noindent
\textbf{(1) 3D annotations are efficient for the in-the-wild scenario.} For in-the-wild images, models trained with paired 3D annotations achieve the best performance. Besides, excluding 80\% paired in-the-wild 3D annotations only increases the reconstruction error by 5\%. When there are no paired in-the-wild 3D annotations existing, incorporating constrained 3D annotations in the training phase can improve the performance and prevent a model from generating unnatural 3D human models. 

\noindent
\textbf{(2) Sparse 2D annotations and dense labeling alone are insufficient.} When there are no paired in-the-wild 3D annotations, using sparse 2D keypoints as the only supervision will decrease the models' performance by 60\%. Besides, using dense labeling as input only brings marginal performance gain.

\noindent
\textbf{(3) Dense correspondence such as IUV map is an effective substitute for 3D annotations.}
After a simple refinement step that removes noisy predictions, dense keypoints sampled from IUV maps can serve as a strong supervision. 
IUV map itself can also serve as a complementary input.
Incorporating dense correspondence can further improve the models' performance by 2.9\% 
or help the model trained with only 20\% paired 3D annotations achieve similar performance of the model trained with full 3D annotations.
Especially, when there are no paired 3D annotations available for in-the-wild images, the model using dense correspondence as supervision can achieve 92\% 
of the performance of an upper-bound models that are trained with a full set of paired 3D in-the-wild annotations.

The contributions of our work are two-fold: 1) We systematically study the effectiveness of different annotations for in-the-wild 3D human recovery. We observe that while using paired 3D annotations leads to optimal results, it is not necessary for 3D human recovery, especially when considering its high annotating cost. 
2) We reveal the effectiveness of incorporating dense correspondence into in-the-wild 3D human recovery. Our experiments show that when there are no in-the-wild annotations available, models trained with dense correspondence can still achieve the same performance as the models trained with 60\% paired in-the-wild 3D annotations. The resulted model can serve as a strong and solid baseline for future studies.


\section{Related Work}

\begin{figure*}[t]
	\begin{center}
		\includegraphics[width=0.85\linewidth]{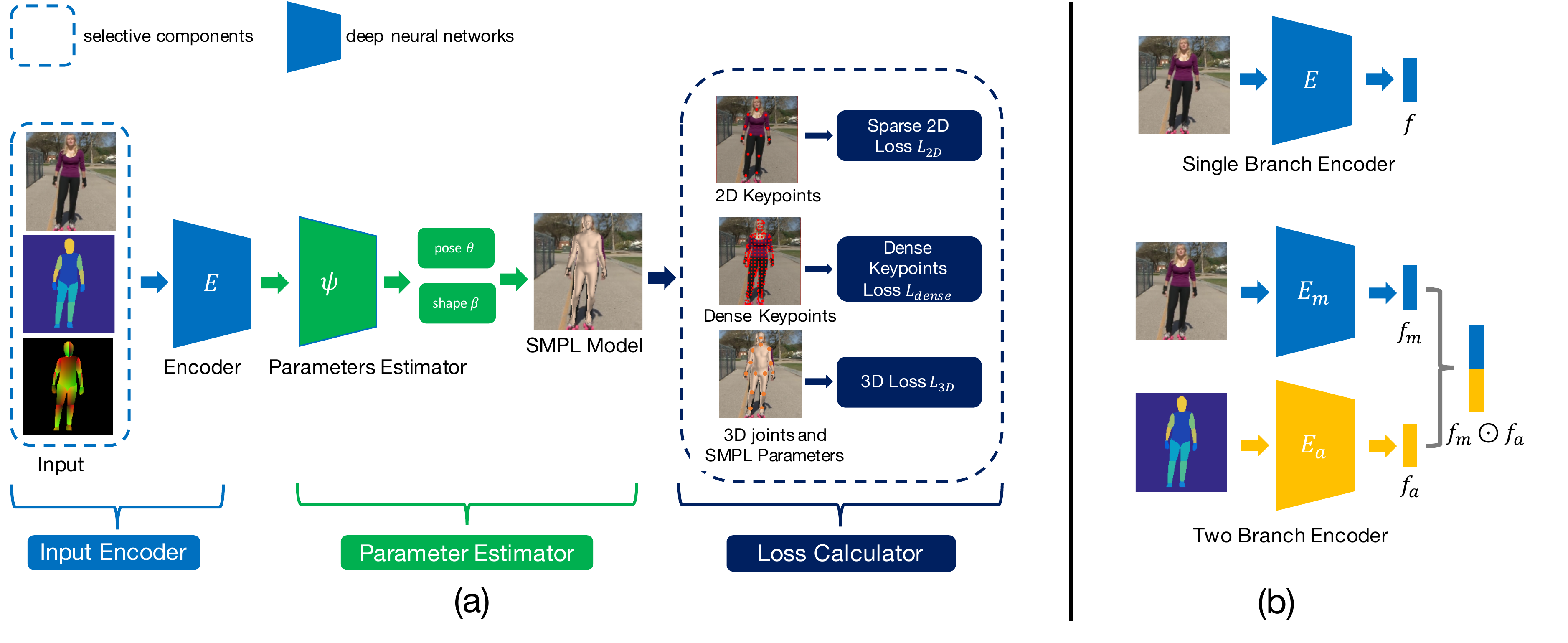}
	\end{center}
	\vskip -0.5cm
	\caption{\textbf{Visualization of the overall framework.} Figure (a) illustrates the overall framework. It is composed of three components. 1) Input encoding part takes inputs and outputs encoded features. 2) Parameter estimator estimates the pose and shape parameters of the SMPL model given the outputs of the encoder. 3) Given estimated parameters, SMPL model generates predicted 3D joints, 2D keypoints and dense keypoints to calculate loss. Figure (b) shows two possible architectures of the input encoder. Input encoder could either be composed of a single branch that only takes one kind of inputs or two branches that takes original images and the other auxiliary inputs.}
	\label{fig:pipeline}
	\vspace{-0.3cm}
\end{figure*}

Recent studies on 3D human recovery mainly use a parametric model - SMPL~\cite{loper2015smpl} to represent human in 3D space.
These studies can be divided into two groups: optimization-based methods and learning-based methods.
Early works are mainly the optimization-based approach.
Bogo~\etal~\cite{bogo2016keep} propose to estimate parameters of SMPL through aligning the predicted models with 2D keypoints. 
Lassner~\etal~\cite{lassner2017unite} extend the algorithm by adding the silhouettes matching loss and 91 landmarks.
Tan~\etal~\cite{tan2017indirect} propose an encoder-decoder architecture, in which the encoder predicts SMPL parameters from images and decoder predicts silhouettes from SMPL. The model is trained with heatmaps of silhouettes. BodyNet~\cite{varol2018bodynet} proposes to predict volumetric 3D human first and then regress SMPL parameters from the predicted volumetric result.

Other recent works \cite{kanazawa2018end,omran2018neural,pavlakos2018learning} share similar pipelines. They all design a CNN-based model to predict the parameters of SMPL. The models are trained with images that come with 2D annotations (2D keypoints) and 3D annotations (3D joints or ground-truth SMPL parameters).
Kanazawa~\etal~\cite{kanazawa2018end} add adversarial loss~\cite{goodfellow2014generative} to judge whether the generated 3D human models are real or not.
Pavlakos~\etal~\cite{pavlakos2018learning} propose to first predict the silhouette and 2D keypoints heatmaps and then use them as the input for the SMPL parameters estimator. Omran~\etal~\cite{omran2018neural} argue that using body part segmentations to replace 2D images as input will enhance the performance of the model.

Most existing studies do not comprehensively investigate the efficiency of each annotation they use.
The other works such as NBF~\cite{omran2018neural} and HMR~\cite{kanazawa2018end} have not completely evaluated the quality of generated 3D models. Their evaluation metrics are partial. Specifically, NBF~\cite{omran2018neural} only evaluates the quality of predicted 3D poses, omitting the predicted shape. HMR~\cite{kanazawa2018end} evaluates in-the-wild images using the accuracy of body part segmentation, which is only a 2D metric.
In order to thoroughly evaluate how models' performance is affected by different factors, in this work, we conduct a series of experiments under a unified framework and training strategy. Besides,  we use the Euclidean distance between predicted and ground-truth 3D meshes as the evaluation metric, which can faithfully reveal the quality of both pose and shape.


\section{3D Recovery with Hybrid Annotations}

To evaluate the efficiency of different annotations for in-the-wild 3D human recovery, we conduct a series of experiments based on a unified framework and train-validation setting. In this section, we first introduce the framework used in the experiments. Then we describe five annotations investigated in this work. Finally, we discuss how to exploit the dense correspondence.

\noindent\textbf{3D Human Model.}
\label{subsec:revisit}
Skinned Multi-Person Linear Model (SMPL)~\cite{loper2015smpl} is a 3D human body model parameterized by the pose and shape parameters.
The shape parameters $\boldsymbol{\beta} \in \mathbb{R}^{10}$ are the first 10 coefficients of PCA components of shape space. Pose parameters $\boldsymbol{\theta} \in \mathbb{R}^{3 \times K}$ represent the 3D rotations for $K = 23$ joints.
%
In general, to specify a complete SMPL model, $(23+1) \times 3 = 72$ pose parameters (three more parameters for global rotation) and $10$ shape parameters are required.

\noindent\textbf{Framework.}
The overall framework, as shown in Figure~\ref{fig:pipeline}, is composed of three components: 1) input encoder 2) parameter estimator 3) loss calculator.
The input encoder $E$ has two variations of architectures: single branch and two branch. 
A two-branch encoder is composed of a main encoder $E_m$ and an auxiliary encoder $E_a$. The main encoder takes images as input while the auxiliary encoder takes one auxiliary input that can either be body part segmentation or IUV maps. The generated main features $f_m$ and $f_a$ are then concatenated to produce the final feature vector $f = f_m \odot f_a $. 
The single branch encoder has only one main branch $E_m$ whose inputs are one category of original images, body part segmentation and IUV maps. It takes inputs and outputs encoded features $f_m$. For single branch encoder, $f = f_m$.

Given encoded feature vectors, the parameter estimator $\boldsymbol{\psi}$, which is composed of two fully-connected layers, predicts the pose and shape parameters of SMPL. The SMPL model then generates the final 3D meshes. 
%
Follow the practice in previous works~\cite{oberweger2015training,carreira2016human,kanazawa2018end}, 
the parameter estimator outputs the residual of parameters $\Delta\Theta$. The final parameters are then obtained by adding the residual with the mean parameters $\bar{\Theta}$. This strategy helps the model to focus on the variance of different images and thus leads to faster convergence.
The parameter estimation process is formulated as follows: $ \Theta = \bar{\Theta} + \psi{(E(I))}$, where $I$ denotes inputs.

In the training phase, the loss calculator further regresses predicted 3D joints, 2D keypoints and dense keypoints obtained from SMPL vertices. The corresponding losses are then calculated using the ground-truth annotations.

\subsection{Hybrid Annotations}

\begin{table}[t] \centering  \footnotesize %
	\caption{\textbf{Role of each annotation.}. The role of different annotations in our experiments.}
	\setlength\tabcolsep{2pt} 
	\begin{tabular}{c|c|c|c|c|c}
		\hline
		Annotation 		& Sparse & Dense     &  Dense          & Constrained  & In-the-wild \\ 
				   		&  2D 	 & Labeling  & Correspondence  &   3D         & 3D          \\
	    \hline
	    Input      		&        & \cmark    &   \cmark        &              &             \\
	    \hline
	    Supervision     & \cmark &           &   \cmark        & \cmark       & \cmark      \\
	    \hline
	\end{tabular}	
	\label{tab:annotation_role}
    \vspace {-0.5cm}
\end{table}

\label{subsec:annotations}
In this section, we discuss different annotations investigated in this work. The annotations include constrained and in-the-wild 3D annotations, sparse 2D annotations, dense labeling and dense correspondence. Depending on the nature of each annotation, they can serve as either input or supervision or both. The role of each annotation in our experiments is listed in Table~\ref{tab:annotation_role}.

\noindent
\textbf{3D Annotations.} 
3D annotations can be divided into two categories according to whether the images are captured in constrained environments or in the wild. Since this paper mainly focuses on in-the-wild scenarios, constrained annotations are mainly used for pre-training. It will also take part in training when there are no paired in-the-wild 3D annotations available. In the loss calculating phase,  for images with ground-truth SMPL parameters, we minimize the distance between predicted and ground-truth parameters. For numerical stability, each pose parameter $\theta_i$ is converted into a $3\times3$ rotation matrix using the Rodrigues formula~\cite{loper2015smpl}. For images with 3D joints annotation, we further minimize the distance between predicted and ground-truth 3D joints. 3D Loss $L_{3D}$ is defined as follows:
\begin{equation}
\label{eq:3d_loss}
\begin{aligned}
& L_{3D\_joints} =  \sum\nolimits_{i=1}^{M}{||(J_{i}^{3D} - \hat{J}_{i}^{3D})||_{2}}, \\        
& L_{SMPL} = \sum\nolimits_{i=1}^{O}{||R(\theta_{i})-R(\hat{\theta}_{i})||_{2} + ||\beta_{i}-\hat{\beta}_{i} ||_{2}}, \\
& L_{3D} = L_{3D\_joints} +  L_{SMPL},
\end{aligned}
\end{equation} 
where $[\theta_{i}, \beta_{i}]$ and $[\hat{\theta_{i}}, \hat{\beta_{i}}]$ are the predicted and ground-truth SMPL parameters, respectively. $M$ and $O$ represent the number of images with 3D joints annotation and ground-truth SMPL parameters. $R: \mathbb{R}^3 \rightarrow \mathbb{R}^{3 \times 3}$ represents the Rodrigues formula.


\noindent
\textbf{Sparse 2D Annotations.}  
To estimate 2D keypoints, the parameter estimator predicts three additional parameters to model the camera $C \in \mathbb{R}^{3}$, two parameters for the camera translation and one parameter for the focal length. $C$ is then used to project the predicted 3D joints $\hat{J}^{3D}$ to 2D keypoints $\hat{J}^{2D}$. The sparse 2D loss $L_{2D}$ can then be defined as:
\begin{equation}
\label{eq:2d_loss}
\begin{aligned}
L_{2D} = \sum\nolimits_{i=1}^{S}{||(J_{i}^{2D} - \hat{J}_{i}^{2D}) \times \mu_{i}||_{1}},\\
\end{aligned}
\end{equation}
where $S$ is the number of training data with 2D keypoints annotation. 
$J_{i}^{2D}$ and $\hat{J}_{i}^{2D}$ denote the predicted and ground-truth 2D keypoints for the $i$th data sample, respectively.
$\mu_{i}$ represent the visibility vectors, where $\mu_{ij} = 1$ means the $j$-th joint of $i$-th sample is visible, otherwise $\mu_{ij} = 0$.

\noindent
\textbf{Dense Labeling.}
Dense labeling investigated in this work is body part segmentation. 
In this work, dense labeling is only used as input. It can either be the sole input or serve as the auxiliary input. In our experiments, body part segmentation is not used as supervision, since the process of obtaining body part segmentation from SMPL predictions is not differentiable.

\noindent
\textbf{Dense Correspondence.}
Our work is in parallel with HoloPose~\cite{guler2019holopose} to incorporate dense correspondence into 3D human reconstruction. We exploit DensePose~\cite{densepose,neverova2019slim}, which establishes dense correspondence between RGB images and human bodies.
Each pixel on a given image can be assigned with a $(I, U, V)$ coordinate, which indicates a specific position on the surface-based human body. $I \in \mathbb{Z}$ indicates which body part this point belongs to and $(U,V) \in \mathbb{R}^2$ is the coordinate of the precise location on the unrolled surface of the body part specified by $I$.

\begin{figure}[t]
	\begin{center}
		\includegraphics[width=1.0\linewidth]{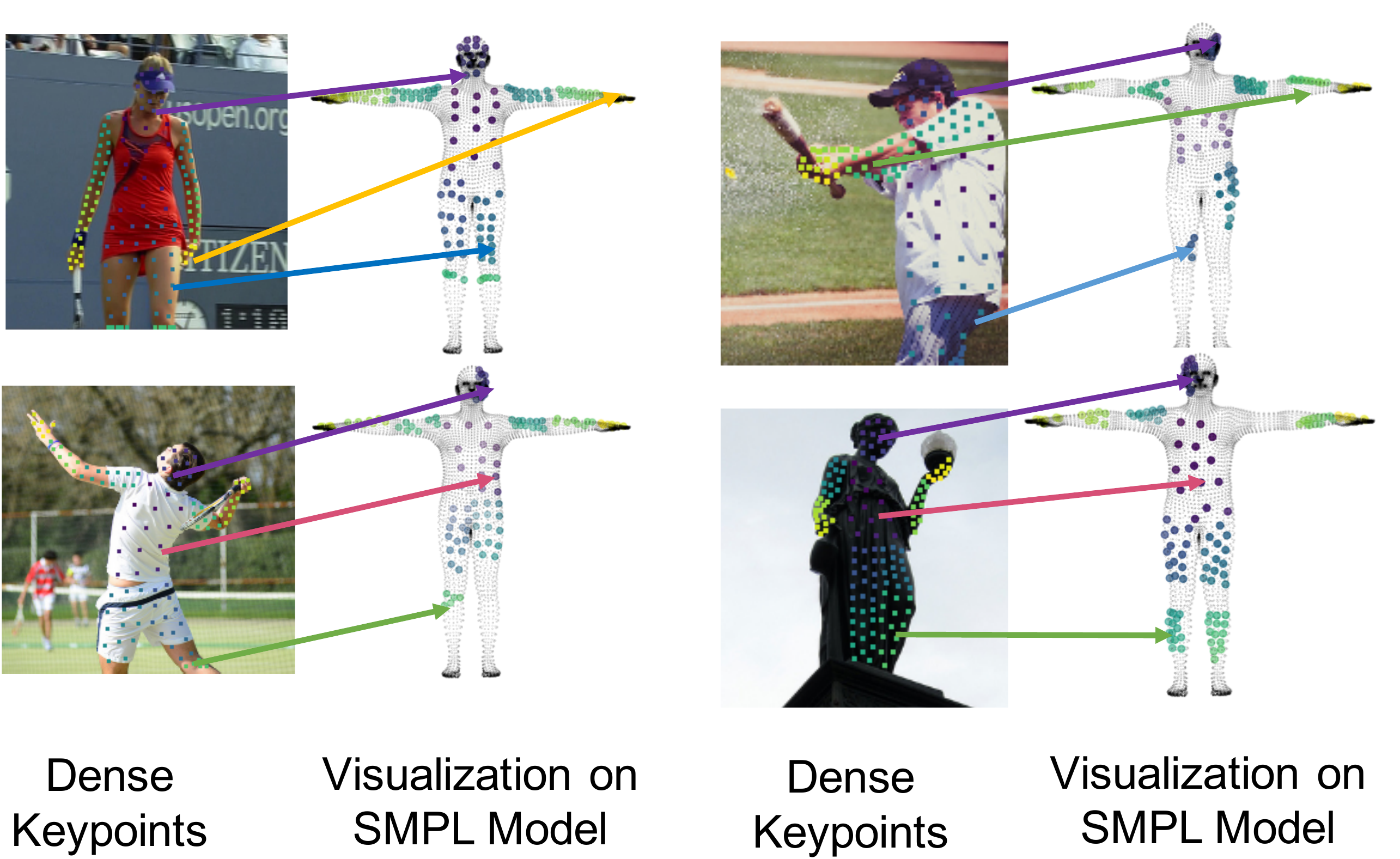}
	\end{center}
	\vskip -0.1cm
	\caption{\textbf{Relationship between DensePose and SMPL.} Corresponding keypoints are annotated with same color.}
	\label{fig:dp_smpl}
	\vspace{-0.1cm}
\end{figure}

There is a close connection between SMPL and IUV in that each vertex of the SMPL model can be assigned an $(I, U, V)$ coordinate. 
In this way, for each point annotated with $(I, U, V)$, we calculate which triangle face of SMPL this point belongs to and the distances from this point to each vertex of the triangle face. These distances form the barycentric coordinates specific to this triangle face. 
Consequently, we have a mapping function $\phi$ that can map the points annotated with $(I,U,V)$ to the vertices of SMPL model. The mapping is provided in the following equation:
\begin{equation}
\label{eq:iuv_to_smpl}
\begin{aligned}
& [v_1, v_2, v_3], [b_1, b_2, b_3] = \phi(I, U, V), \\
\end{aligned}
\end{equation}
where $v_i$ denotes the index of selected vertices and $b_i$ represent the barycentric coordinate. 
We show some examples in Figure~\ref{fig:dp_smpl} to demonstrate the relationship between DensePose model and SMPL.

In the training phase, IUV maps generated by DensePose can either be used as inputs or used for providing supervision. When serving as supervision, dense keypoints are sampled from IUV maps and used to calculate dense keypoint loss. Each dense keypoint is composed of two parts: the coordinate $(x,y)$ on the RGB images and the coordinate $(I,U,V)$. For simplicity of notation, we denote $(I,U,V)$ coordinate as $D$. 
Given $D$, Equation~\eqref{eq:iuv_to_smpl} is used to calculate which vertices $\mathbf{f}=[v_{1}, v_{2}, v_{3}]$ this point is closest to and the corresponding barycentric coordinates $\mathbf{b} = [b_{1}, b_{2}, b_{3}]$.  
After obtaining $\mathbf{f}$ and $\mathbf{b}$, we project predicted SMPL vertices $\hat{P} \in \mathbb{R}^{3\times N}$ to 2D space $\hat{P}^{2D} \in \mathbb{R}^{2\times N}$ using the similar method of projecting 3D joints to 2D keypoints.
Finally, we can obtain the predicted dense keypoints by weighted averaging the selected 2D vertices using barycentric coordinates and calculate the dense keypoint loss between the pixel coordinates of predicted and ground-truth dense keypoints. The whole process is formulated as:
\begin{equation}
\label{eq:dense_loss}
\begin{aligned}
	& [v_{i1}, v_{i2}, v_{i3}], [b_{i1}, b_{i2}, b_{i3}] = \phi(D_{i}), \\
    & \hat{X_{i}} = \sum\nolimits_{j=1}^{3}{\hat{P}^{2D}_{i}[v_{ij}] \times b_{ij}}, \\
    &  L_{dense} = \sum\nolimits_{i=1}^{T}{||(X_{i}- \hat{X}_{i})||_{1}},
\end{aligned}
\end{equation}
where $T$ is the number of images with dense keypoints annotations, $\phi: \mathbb{Z}\times\mathbb{R}^{2} \rightarrow \mathbb{Z}^{3} \times \mathbb{R}^3$ is the mapping function defined in Equation~\eqref{eq:iuv_to_smpl}.

\subsection{Learning}
\label{sec:learning}
\noindent
\textbf{Sampling Strategy for Dense Correspondence.}
The dense points drawn from IUV maps cannot be employed directly since they frequently contain wrong predictions. For example, the left foot could be wrongly predicted as the right foot.
To avoid erroneous points corrupting our model, we perform refinement by using accurate sparse keypoints as the reference.
For each visible 2D keypoint, we check the values of IUV map in the $3 \times 3$ grid centering at it and select the value of `I' (which indicates body part) that appears most frequently as the body part prediction of IUV map surrounding this keypoint. Then we check whether the body part prediction matches the 2D keypoint or not.

After finding the erroneous region, our sampling scheme sets the IUV map of this sub-area to be background in a recursive manner: 
We first set the IUV value of the keypoint to be background, then we check the $3 \times 3$ grid around it and determine the pixels whose value of `I' equals to the \textit{surrounding IUV} and set their IUV values to be background.
Further, we check the $3 \times 3$ grids centering at these pixels and determine more pixels using the same condition. The process is conducted recursively until there are no more pixels found.
The above process is conducted on each keypoint to refine the whole IUV map before we use the map as input and for sampling dense keypoints. A more detailed description along with an illustration figure can be found in the appendix~\ref{lab:sample_dense_keypoints}.

\noindent
\textbf{Overall Loss Function.}
The overall loss $L$ is defined as:
\begin{equation}
	\label{eq:overall_loss}
		\begin{aligned}
			L = \lambda_{1}L_{3D} + \lambda_{2}L_{2D} + \lambda_{3}L_{dense}.
		\end{aligned}
\end{equation}
Detail values of $\lambda$ used in the experiments is listed in the appendix~\ref{lab:more_implementation_details}.


\section{Experiments}

\begin{table}[t] \centering  \small
	\caption{\textbf{FLOPs and model size of different architectures.}}
	\vskip 0.1cm
	\begin{tabular}{c|c|c}
		\hline
		Encoder             	    &   FLOPs $\times 10^9$        & Model Size (mb)  \\
		\hline
		ResNet-101                  &   7.803                      & 174.97 \\
		ResNet-50                   &   4.090                      & 102.27 \\
		\hline
		ResNet-50 \& ResNet-18      &   5.905                      & 150.97 \\
		ResNet-18 \& ResNet-18      &   3.630                      & 97.783 \\ 
		\hline
	\end{tabular}	
	\label{tab:flops_and_size}
	\vspace{-0.4cm}
\end{table}

\begin{table*}[t] \centering  \small %
	\caption{\textbf{Influence of different annotations.} The evaluation metrics are PVE, MPJPE and PVE-T, separately. For all metrics, lower is better. ``3D'' refers to paired in-the-wild 3D annotations. ``20\% 3D'' refers to 20\% randomly selected 3D annotations. ``Sparse 2D'' refers to sparse 2D keypoints. ``Dense'' refers to dense correspondence, namely, IUV maps generated by DensePose~\cite{densepose,neverova2019slim}.  }
	\vskip 0.1cm
	\setlength\tabcolsep{2pt}
	\begin{tabular}{c|c|c|c|c|c}
		\hline
		Supervision $\rightarrow$  	& 	3D \& Dense \&							& 20\% 3D \& Dense \&	   & \multirow{2}{*}{3D \& Sparse 2D}   &\multirow{2}{*}{Dense \& Sparse 2D}     	&\multirow{2}{*}{Sparse 2D Only} \\	
		Input $\downarrow$			& 	 Sparse 2D								&  Sparse 2D			   & 		      						&    	 									&	  							\\
		\hline
		IUV Only                	& \textbf{120.0} / \textbf{103.1} / 31.8 	& 125.0 / 107.2 / 32.6	   &  125.2 / 106.4 / 32.1	   			&  138.7 / 121.2 / 54.7    					&	 204.3 / 177.0 / 92.1	   \\
		Segment Only                & 123.0 / 105.1 / 32.7 						& 126.7 / 110.0 / 33.2	   &  124.8/ 107.8 / 31.7      			&  147.4 /  130.1  / 55.9   				&    203.8 / 176.7 / 93.3   \\
		Image Only					& 123.7 / 105.9 / 30.9 						& 127.5 / 110.6 / 32.2	   &  127.4 / 108.5 / 30.7     			&  137.7 / 120.3 / 51.7						&	 203.2 / 178.5 / 106.2    \\
		\hline
		Image \& IUV				& 122.4 / 105.1 / \textbf{30.2} 			& 125.0 / 107.6 / 32.1	   &  125.5 / 107.3 / 30.7	    		&  133.8 / 117.2 / 52.5     				&    197.3 / 172.8 / 107.9    \\
		Image \& Segment			& 121.5 / 104.3 / 31.0						& 126.4 / 107.0 / 31.6	   &  125.8 / 106.8 / 31.5	    		&  142.2 / 124.2 / 56.6						&    201.2 / 177.5 / 101.7    \\
		\hline
	\end{tabular}	
	\label{tab:supervision}
	\vspace{-0.4cm}
\end{table*}

We first introduce the datasets and evaluation metrics used in this work. 
In our experiments, we employ four datasets: Human3.6M~\cite{ionescu2014human3}, COCO-DensePose~\cite{densepose}, UP-3D~\cite{lassner2017unite} and 3DPW~\cite{vonMarcard2018}. Experiments are mainly conducted on UP-3D dataset since it is the only in-the-wild dataset with SMPL annotations.
We compare our methods with previous state-of-the-arts on UP-3D, 3DPW and COCO-DensePose datasets.

\noindent
\textbf{Human3.6M.} Human3.6M~\cite{ionescu2014human3} is an indoor dataset. Following HMR~\cite{pavlakos2018learning}, we use Mosh~\cite{loper2014mosh} to collect ground-truth SMPL parameters from raw 3D Mocap markers. In our experiment, the data of Human3.6M is used in pre-training. It is also used in training when there are no paired in-the-wild 3D annotations available.

\noindent
\textbf{COCO-DensePose.} COCO-DensePose dataset~\cite{densepose} is a newly released dataset that builds dense correspondence between images and body part surface. Images in this dataset are all selected from the keypoints MS-COCO dataset~\cite{lin2014microsoft}. Researchers in~\cite{densepose} re-annotate each selected image with about $100 \sim 150$ dense keypoints. We train our model on the training set and test the models on the evaluation set. 

\noindent
\textbf{UP-3D.} This dataset is built by Lassner~\etal~\cite{lassner2017unite}. They pick images from four pose estimation datasets including: LSP~\cite{Johnson10}, LSP-extened~\cite{Johnson11}, MPII~\cite{andriluka20142d} and FashionPose~\cite{dantone2014body}. The researchers extend SMPLify~\cite{bogo2016keep} and fit the model to those images. Then they ask human annotators to pick the samples with good fitness. 

\noindent
\textbf{3DPW.} This dataset is built by Von~\etal~\cite{vonMarcard2018}. They estimate 3D poses using a single hand-held camera and a set of IMUs attached at body limbs. 3D body shapes are obtained through 3D scans. This dataset cannot be counted as a totally in-the-wild dataset since the data are collected by several actors performing different actions. We compare our methods with previous state-of-the-arts, \eg, HMR~\cite{kanazawa2018end}. 

\phantomsection
\label{para:evaluation_metrics}
\noindent
\textbf{Evaluation Metrics.}
For COCO-DensePose dataset, the evaluation metric is the dense keypoints distance introduced in Equation~\eqref{eq:dense_loss}. It is abbreviated as DKD in the following sections.
For other datasets with SMPL annotations, we use the mean per-vertex error (PVE) proposed by Pavlakos~\etal~\cite{pavlakos2018learning} as the metric, which computes the Euclidean distance between ground-truth SMPL vertices and the predicted SMPL vertices. We also report mean per joint position error (MPJPE) on SMPL joints to reveal the quality of pose recovery and PVE between SMPL vertices whose shape parameters come from ground-truth and prediction while pose parameters are set to be the same (in the experiment, pose parameters are all set to be zero). We use this metric to reveal the quality of shape recovery and abbreviate it as PVE-T, where ``T'' refers to T-pose.

\noindent
\textbf{Implementation Details}.
All images are cropped according to the bounding boxes of humans. These images are further padded and scaled to $224 \times 224 $. During training, images are randomly flipped and scaled for data augmentation.
As depicted in Figure~\ref{fig:pipeline}, the input encoder has two architectures. In most experiments, the single branch encoder is based on ResNet-101~\cite{he2016deep} while the main encoder and auxiliary encoder of the two branch architecture are based on ResNet-50 and ResNet-18, separately.
In this way, models with different architectures have comparable FLOPs and model size. The overall FLOPs and size of models adopting different input encoders are listed in Table~\ref{tab:flops_and_size}.

We assign additional fully-connected layers at the top of the input encoder to map the feature vectors to $85$ dimensions.  The final output vectors contain pose parameters $\boldsymbol{\theta}$ ($72$ dimensions), shape parameters $\boldsymbol{\beta}$ ($10$ dimensions) and camera model $C$ ($3$ dimensions).

\subsection{The Effectiveness of Hybrid Annotations}
\label{sec:hybrid_annotation}

In this subsection, we study the efficiency of different annotations when serving as inputs or supervisions. In all the experiments, sparse 2D keypoints are always assumed to be available, as annotating 2D keypoints is quite cheap. Alternatively, precise results can be obtained using state-of-the-arts 2D pose estimation algorithms~\cite{xiao2018simple,cao2017realtime}.
For each input type, we adopt five different supervision combinations, including 3D annotations, 3D annotations plus dense correspondence, randomly selected 20\% 3D annotations plus dense correspondence, dense correspondence only and sparse 2D keypoints only. The results are listed in Table~\ref{tab:supervision}.

\begin{figure*}[t]
	\begin{center}
		\includegraphics[width=0.95\linewidth]{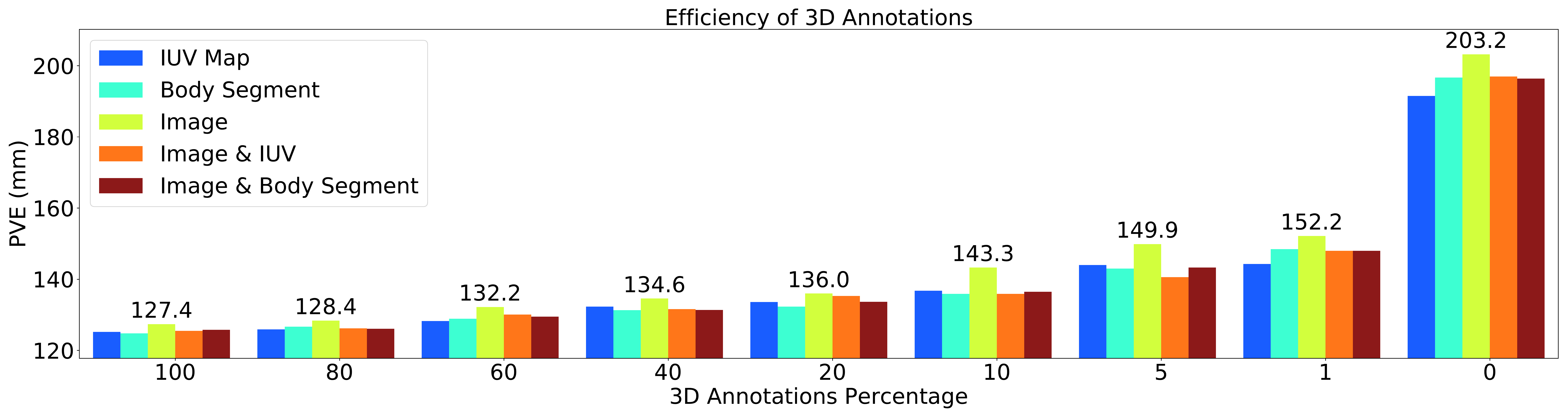}
	\end{center}
	\vskip -0.4cm
	\caption{\textbf{Influence of 3D annotations.} We test different models on the test set of UP-3D~\cite{lassner2017unite} using the per-vertex error (abbreviated as PVE, the unit is mm.) as the metric. The figure shows that 3D annotations are very efficient.} 
	\label{fig:3D_annotation}
	\vskip -0.2cm
\end{figure*}

\noindent
\textbf{Influence of Supervision.}
Detailed numbers in this subsection are calculated by comparing the models that take images as the only input (the fourth row of Table~\ref{tab:supervision}). Same conclusions can be drawn from other models that use different inputs.
It is not surprising that 3D annotations can provide the best guidance for in-the-wild 3D human recovery while sparse 2D keypoints are not as efficient. 
Dense correspondence, namely, IUV maps generated by DensePose~\cite{densepose,neverova2019slim}, is an effective annotations for in-the-wild 3D human recovery.
The model trained with sampled dense keypoints and sparse 2D keypoints can achieve the 92\% performance of the model trained with full set 3D annotations.
%
%
Furthermore, the model trained with hybrid of only 20\% 3D annotations and dense correspondence achieve comparable performance with the model trained with full 3D annotations. 
%
Besides, the performance of the model trained with full 3D annotations can be improved by 2.9\% through incorporating dense correspondence into training.

\noindent
\textbf{Influence of Input.}
%
Five input combinations are exploited in our experiments, including 1) images only, 2) IUV maps only, 3) body part segmentation only, 4) images plus IUV maps, 5) images plus body part segmentation. The first three categories adopt a single branch architecture and the last two use the two-branch architecture. For a fair comparison, IUV maps and body part segmentations are both generated by DensePose~\cite{densepose} model.
Experimental results in Table~\ref{tab:supervision} show that when sparse 2D keypoints serve as the only supervision, incorporating auxiliary inputs including body part segmentation or IUV maps can only improve the models performance by 1.5\% in average. 
It is marginal when compared with 32\% improvement brought by incorporating sampled dense keypoints from IUV maps into supervision while still using the images as the only input.

%

\begin{table}[t] \centering  \small %
	\caption{\textbf{Influence of pose and shape parameters}. The evaluation metrics are: PVE, MPJPE and PVE-T, separately.}
	\vskip 0 cm
	\setlength\tabcolsep{1pt}
	\begin{tabular}{c|c|c}
		\hline
		3D Loss  $\rightarrow$  			&   \multirow{2}{*}{3D Pose Only}     &  \multirow{2}{*}{Shape parameters Only}    \\ 
		Other Supervision $\downarrow$		&	        					      &    	 	       \\
		\hline
		DC \& Sparse 2D             		&    131.3 / 116.6 / 59.0 		      &  148.5 / 127.3 / 30.6	                    \\
		Sparse 2D Only              		&    164.0 / 148.2 / 117.0    	      &  220.0 / 180.6 / 31.4            			\\
		\hline
	\end{tabular}	
	\label{tab:seperate_params}
	\vspace{-0.5cm}
\end{table}

\subsection{Exploit 3D Annotations} 
\noindent
\textbf{Influence of Separate Parameters.}
We separately evaluate the influence of SMPL pose and shape parameters by using only one of them during training. The results shown in Table~\ref{tab:seperate_params} suggest that: (1) 3D poses and SMPL parameters explicitly affect MPJPE and PVE-T, respectively. (2) 3D poses have more influence on the model's overall performance. 
Besides, the results in Table~\ref{tab:supervision} show that when both pose and shape parameters are used in training, MPJPE and PVE-T are nearly consistent with the PVE. Therefore, we only report PVE in the following experiments.

\noindent
\textbf{Efficiency of 3D Annotations.}
We then evaluate the effectiveness of in-the-wild 3D annotations. Models in this section are all trained with 3D annotations and sparse 2D annotations. 
In these experiments, the number of paired 3D annotations is reduced gradually from 100\% to 0\% (0\% means only using sparse 2D annotations in training). The results are shown in Figure~\ref{fig:3D_annotation}. We only show detailed results of models taking images as the only inputs. Detailed experiment results of all the models can be found in the appendix~\ref{lab:more_3d}.
From the Figure~\ref{fig:3D_annotation}, we find that 3D annotations are efficient. For instance, the reconstruction error only increases by 6\% when 80\% 3D annotations are excluded from training. 
On the contrary, sparse 2D annotations are incompetent in guiding 3D human recovery. When there are no paired 3D data available, the performance drops drastically. The reconstruction error is 34\% larger than models trained with only 1\% of paired 3D data.

\begin{figure}[t]
	\begin{center}
		\includegraphics[width=1.0\linewidth]{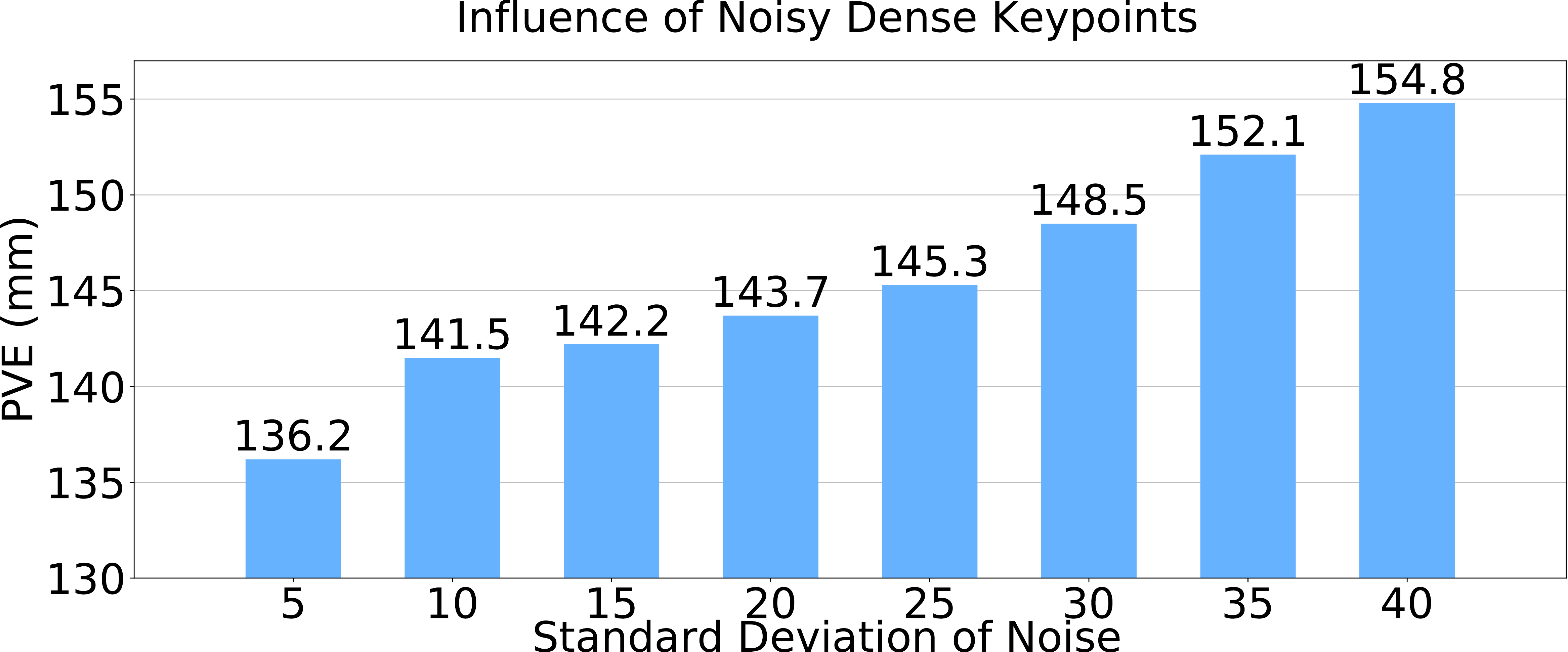}
	\end{center}
	\vskip -0.3cm	
	\caption{\textbf{Influence of noisy dense correspondence}. In this experiment, we add Gaussian noise to IUV maps. The mean ($\mu$) is fixed to be $0$ and standard deviation varies from $5$ to $40$.} 
	\label{fig:dp_noise}
	\vspace{-0.3cm}
\end{figure}

\begin{figure}[t]
	\begin{center}
		\includegraphics[width=1.0\linewidth]{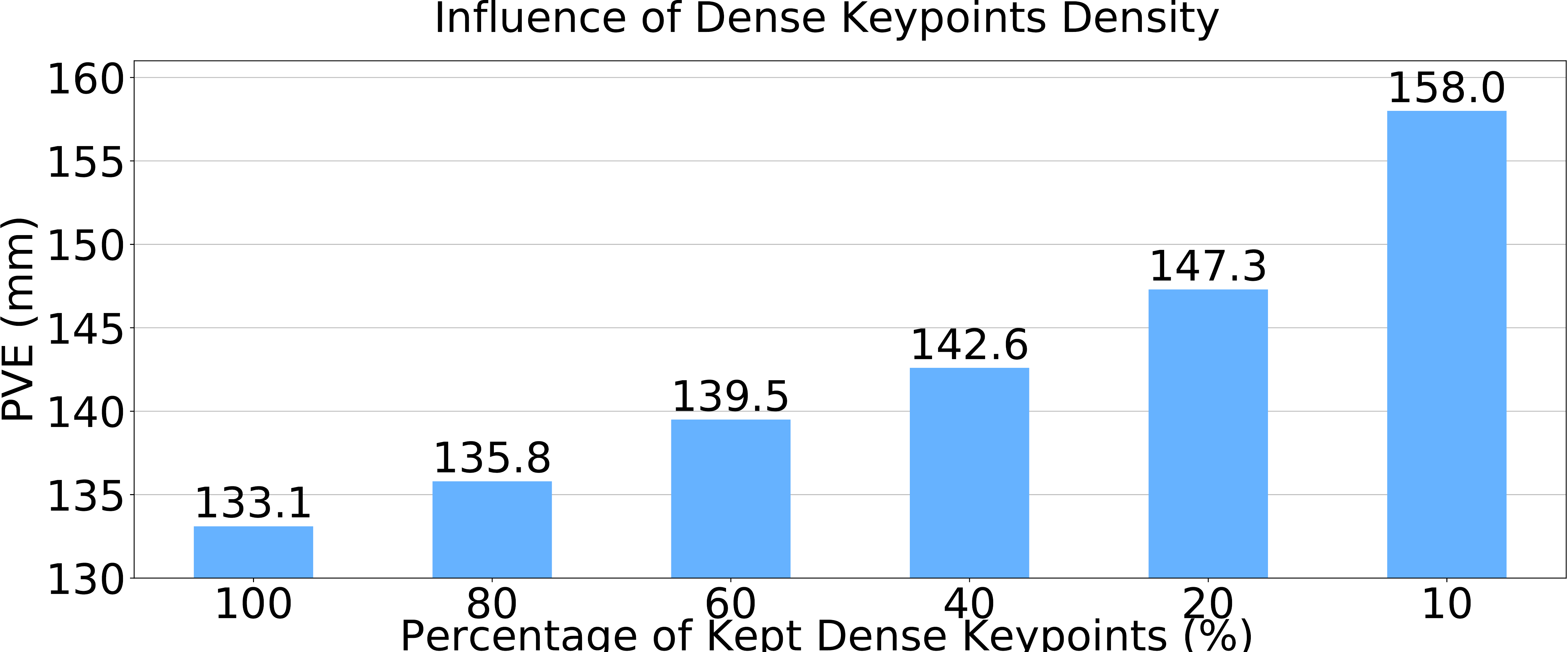}
	\end{center}
	\vskip -0.3cm	
	\caption{\textbf{Influence of dense keypoints density} In this experiment, sampled dense keypoints are randomly discarded. Performance of the model drops gracefully when more than 60\% keypoints are retained. Even with only $10 \sim 15$ dense keypoints kept, they are still significantly more efficient than sparse 2D keypoints.}
	\label{fig:dp_density}
	\vspace{-0.5cm}
\end{figure}

\begin{figure*}[t]
	\begin{center}
		\includegraphics[width=0.9\linewidth]{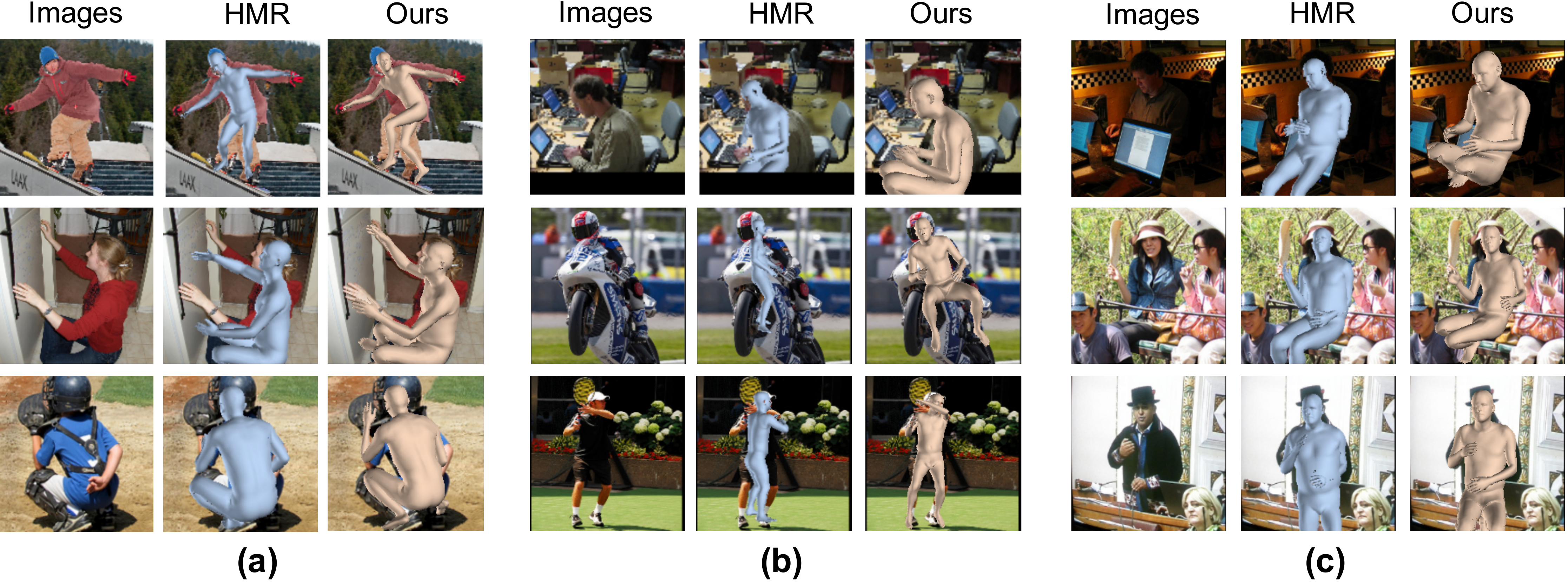}
	\end{center}
	\vskip -0.5cm
	\caption{\textbf{A comparison between our model and HMR~\cite{kanazawa2018end}}. ``Our model'' refers to the model that adopts the framework in Figure~\ref{fig:pipeline}. It  uses images and IUV maps as input and it is trained with dense correspondence and sparse 2D keypoints. (a) shows that our model can generate better-aligned results. (b) shows that our model still works well on some tough samples. (c) shows that our model is capable of generating natural results when HMR fails. The images all come from COCO-DensePose dataset~\cite{densepose}.}
	\label{fig:qualitative_compare}
	\vskip -0.3cm
\end{figure*}

\subsection{Exploit Dense Correspondence}
\label{sec:exploit_dense_correspondence}
Inspired by surprising efficiency of dense correspondence as observed in Table~\ref{tab:supervision}, we further investigate its effectiveness in this subsection. Models in this subsection all take images and IUV maps as inputs.

\noindent
\textbf{Influence of Noisy Dense Correspondence.}
As stated before, dense keypoints used as supervision are sampled from IUV maps, which might contain errors. We refine IUV maps as described in section~\ref{sec:learning}. If we directly use raw IUV maps the performance drops by 20.1\%. 
We further study how noise in $U$ and $V$ influence the models' performance, since the refinement process only removes potential errors in $I$.
We add Gaussian noise to $U$ and $V$, whose values lie in $[0, 255]$. The mean ($\mu$) of Gaussian noise is fixed to be $0$ and the standard deviation ($\sigma$) varies from $5$ to $40$. 
The result is illustrated in Figure~\ref{fig:dp_noise}. The results show that our method is robust to the noise. The performance of the model drops gracefully while the variance of noise is less than $10$. Even the variance of noise increases to $40$, using noisy dense keypoints could still enhance the performance of the model considerably.

\noindent
\textbf{Influence of Dense Keypoints Density.}
Each image in the COCO-DensePose dataset is annotated with $100 \sim 150$ dense keypoints. We sample same amount of dense keypoints on UP-3D. In this subsection, we study the influence of dense keypoints density by randomly discarding part of dense keypoints and train the model using the remaining ones. The number of dense keypoints is reduced gradually from 100\% to 0\% ($0$ means using only 2D keypoints.). The results are shown in Figure~\ref{fig:dp_density}. The performance drops gracefully when more than 60\% dense keypoints are retained. Besides, models trained with only $10 \sim 15$ dense keypoints still have significantly higher performance than models trained with only sparse 2D keypoints. Experiment results in this subsection is useful for real-life application in that lots of efforts in annotating dense keypoints could be saved with a little sacrifice in the final performance.

\begin{table}[t] \centering  \footnotesize
	\caption{\textbf{Comparison with state-of-the-art methods}. This table presents the evaluation results on COCO-DensePose dataset~\cite{densepose} (CODP is used for the simplicity of notation.) using DKD (Dense Keypoints Distance), the unit is mm. It also presents evaluation results on UP-3D dataset~\cite{lassner2017unite} and 3DPW dataset~\cite{vonMarcard2018} using PVE (Per-Vertex Error), the unit is mm. For all the metrics, lower is better.  ``Ours-3D'' refers to the proposed model trained using paired 3D annotations. ``Ours-DC'' refers to the proposed model trained using only dense correspondence and sparse 2D annotations. }
	\setlength\tabcolsep{2pt}
	\vskip 0.1cm
	\begin{tabular}{c|c|c|c}
		\hline
		Dataset $\rightarrow$                      & CODP~\cite{densepose} & UP-3D~\cite{lassner2017unite} & 3DPW~\cite{vonMarcard2018} \\
		Metric $\rightarrow$    	               &    DKD   			   &  PVE             			   &   PVE                      \\
		Methods $\downarrow$                       &    (mm)  			   &  (mm)                         &   (mm)					    \\
		\hline\hline
		Lassner~\etal~\cite{lassner2017unite}      &    --    			   &    169.8         			   &     --        			\\
		NBF~\cite{omran2018neural}                 &    --   			   &   134.6         			   &    --                  \\
		HMR~\cite{kanazawa2018end}                 &   102.7 			   &   149.2          		       &    161.0               \\
		Pavlakos~\etal~\cite{pavlakos2018learning} &    --     		       &  \textbf{117.7}               &    --                  \\    
		\hline
		Ours-3D								   	   &	--				   &   	122.2                      & \textbf{152.9}          \\
		Ours-DC                                    & \textbf{51.8}         &    137.5			           &   165.3          		\\
		\hline
	\end{tabular}	
	\label{tab:quantitative_compare}
	\vspace {-0.5cm}
\end{table}

\subsection{Comparison with State-of-the-arts}
\noindent
\textbf{Quantitative Results.}
For UP-3D, we compare our model with both the optimization-based methods~\cite{lassner2017unite} and the learning-based methods~\cite{kanazawa2018end,pavlakos2018learning,omran2018neural}. 
%
For COCO-DensePose, we mainly compare our method with HMR~\cite{kanazawa2018end}, since HMR is the only method that has been trained on COCO~\cite{lin2014microsoft} dataset, which covers all the images in CODP dataset.
For 3DPW, we train HMR on the training set and compare our methods with it on the testing set.
The results are shown in Table~\ref{tab:quantitative_compare}. ``Ours-3D'' refers to the proposed model trained using paired 3D annotations. ``Ours-DC'' refers to the proposed model trained with only dense correspondence and sparse 2D annotations. 
These two models both adopt the two-branch encoder that takes images and IUV maps as input. We use ResNet-18 as the backbone for ``Ours-3D'' and ``Ours-DC'' to guarantee a fair comparison, since models of previous works such as HMR~\cite{pavlakos2018learning} and NBF~\cite{omran2018neural} are all based on ResNet-50~\cite{he2016deep}.  

When 3D data is available, our method surpasses or performs comparably with previous state-of-the arts, demonstrating that our model is simple yet efficient. On UP-3D dataset, it is noteworthy that our model trained using dense correspondence are comparable with most of the previous methods despite no paired in-the-wild 3D annotations are used in training.

\noindent
\textbf{Qualitative Results.} We show some qualitative results of our model and HMR~\cite{kanazawa2018end} in Figure~\ref{fig:qualitative_compare}.
``Our model'' refers to the model that adopts the framework in Figure~\ref{fig:pipeline}, which uses images and IUV maps as input and it is trained with dense correspondence and sparse 2D keypoints.
The observations for each subfigure are given as follows:
(a) shows that our model generates better-aligned and more precise 3D human models than HMR does. 
(b) shows that when HMR fails on images with extreme poses or scales, our model can still generate plausible results.
(c) shows that in some cases HMR generates erroneous 3D models while our method generates more natural results.


\section{Conclusion}

We have performed a systematic study of the cost and efficiency trade-off of hybrid annotations used in in-the-wild 3D human recovery. Through extensive experiments, we find that paired in-the-wild 3D annotations are not irreplaceable as commonly believed. Interestingly, in the absence of paired 3D data, the models that exploits dense correspondence can achieve 92\% of the performance compared to the models trained with paired 3D data. 
We further benchmark against previous state-of-the-art methods on UP-3D~\cite{lassner2017unite} and 3DPW~\cite{vonMarcard2018} dataset. Without paired in-the-wild 3D annotations, the model achieves comparable performance with most of the previous state-of-the-arts methods trained with paired 3D annotations. We demonstrate that dense correspondence is a new supervision form that is promising and competitive for in-the-wild 3D human recovery. Considering its high efficiency and relatively low annotating cost, our models can serve as a strong reference for future research.

\vspace {0.15cm}
\noindent \textbf{Acknowledgements.} This work is partially supported by the Collaborative Research grant from SenseTime Group (CUHK Agreement No. TS1610626 \& No. TS1712093), the General Research Fund of the Hong Kong (CUHK 14209217), Singapore MOE AcRF Tier 1 (M4012082.020), NTU SUG, and NTU NAP.

{\small
\bibliographystyle{ieee_fullname}
\bibliography{short,DCT}
}

\clearpage

\appendix

\section{Sampling Dense Keypoints}
\label {lab:sample_dense_keypoints}
Since dense keypoint annotations are only available in COCO-DensePose dataset and training models purely using sparse 2D keypoints will lead to suboptimal results, we present an effective method for generating dense keypoints for other in-the-wild images that only annotated with sparse 2D keypoints. 
An effective way is to directly sample points from the IUV maps produced by the DensePose model.  

\begin{figure*}[t]
	\begin{center}
		\includegraphics[width=\linewidth]{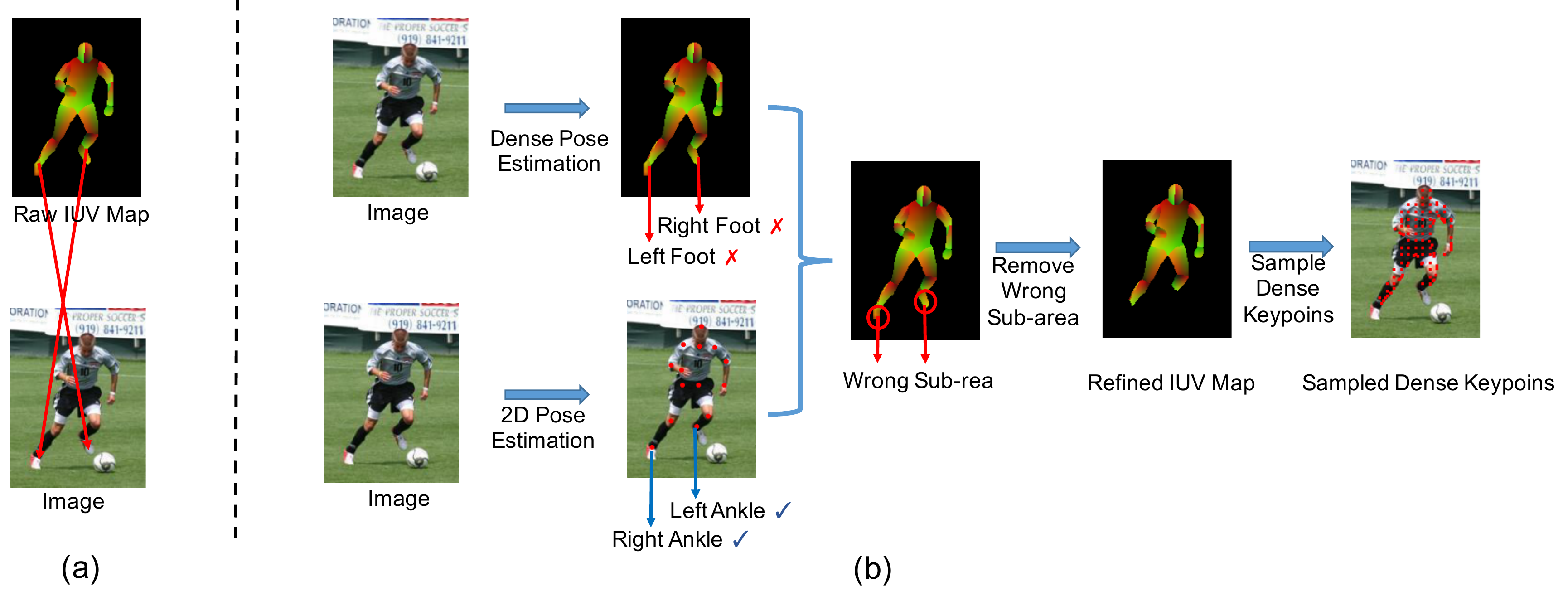}
	\end{center}
	\caption{Figure (a) demonstrates that the raw IUV map might contain errors. Figure (b) shows the process of refining the IUV maps. The generated IUV map is compared with the 2D keypoints. If they are not consistent, \eg,  the sub-area around ``right ankle'' is predicted as ``left foot'', then we discard this sub-area by assigning it as background. We compare each keypoint with the predicted IUV maps surrounding it and remove the inconsistent part.}
	\label{fig:refine_IUV}
\end{figure*}

The dense points drawn from IUV maps cannot be employed directly since the maps frequently contain wrong predictions. As Figure~\ref{fig:refine_IUV} (a) shows, the left foot is wrongly predicted as the right foot while the right foot is predicted as the opposite.
To avoid erroneous points corrupting the learning of our model, we perform sampling of dense points by using accurate sparse keypoints as reference.
Specifically, for each visible 2D keypoint, we check the values of IUV map in the $3 \times 3$ grid centering at it and select the value of `I' (which indicates body part) that appears most frequently as the body part prediction of IUV map surrounding this keypoint. 
Then we chech whether the \textit{surrounding IUV} is consistent with the 2D keypoint.
For example, if a keypoint is labeled as ``right ankle'' but the \textit{surrounding IUV} is ``left foot'', then this sub-area is assigned as erroneous region. 

After finding the erroneous region, our sampling scheme will set the IUV map of this sub-area to be background in a recursive manner: 
We first set the IUV value of the keypoint to be background, then we check the $3 \times 3$ grid around it and determine the pixels whose value of `I' equals to the \textit{surrounding IUV} and set their IUV values to be background.
Further, we check the $3 \times 3$ grids centering at these pixels and determine more pixels using the same condition. The process is conducted recursively until there are no more pixels found.
The above process is conducted on each keypoint to refine the whole IUV map before we use the map as the complementary input and for sampling dense keypoints. The sampling process is depicted in Figure~\ref{fig:refine_IUV} (b).

\section{Efficiency of 3D Annotations.} 
\label {lab:more_3d}
\noindent \textbf{Detailed experiment results}. Detailed experiment results in Figure~\ref{fig:3D_annotation} is listed in Table~\ref{tab:3D_annotation}.
In experiments, the amount of paired 3D annotations used in the training phase is reduced gradually from 100\% to 0\% (0\% means only using sparse 2D annotations in training).
From the table, we find that 3D annotations are quite efficient. The reconstruction error only increases by 6\% when 80\% 3D annotations are excluded from training. 

\noindent \textbf{Influence of constrained 3D.}
We also investigate constrained annotations. The experiment results are listed in Table~\ref{tab:constrained}. When paired in-the-wild 3D annotations exist, using constrained 3D annotations barely brings improvement. However, when there are no paired in-the-wild 3D annotations exist, incorporating constrained 3D annotations into training improves the performance of models bys 30\%.

\section{Implementation Details}
\label {lab:more_implementation_details}
In this section, we discuss more implementation details.
In the training phase, the whole model is first pretraind using 3D data from Human3.6M dataset~\cite{ionescu2014human3}, then it is finetuned on the COCO-DensePose~\cite{densepose}, UP-3D~\cite{lassner2017unite} and 3DPW~\cite{vonMarcard2018}. 
For COCO-DensePose dataset, we train our model with ground truth dense keypoints and 2D keypoints. For UP-3D and 3DPW dataset, our model is trained with the combination of 3D annotations,2D keypoints and sampled dense keypoints. The sampled dense keypoints are obtained based on the method described in Section~\ref{lab:sample_dense_keypoints}.

In the training phase, the batch size is set to 128. Adam optimizer~\cite{kingma2015adam} with $1e-4$ is adopted in the whole training phase. The model gets converged after $40 \sim 50$ epochs. Especially, if all the losses including 3D, dense and 2D are used in training, their balance weights are $10$, $1$, $10$, respectively. If only two losses are used, their balance weights are set to be both $10$.

\clearpage

\begin{table*}[t] \centering
	\caption{\textbf{Influence of 3D annotations.} This table lists detailed experiment results of Figure~\ref{fig:3D_annotation}.}
	\begin{tabular}{|c|c|c|c|c|c|c|c|c|c|}
		\hline
		Kept 3D Annotations (\%) $\rightarrow$  & \multirow{2}{*}{100} & \multirow{2}{*}{80} & \multirow{2}{*}{60} & \multirow{2}{*}{40} & \multirow{2}{*}{20} & \multirow{2}{*}{10} & \multirow{2}{*}{5} & \multirow{2}{*}{1} & \multirow{2}{*}{0} \\
		Input $\downarrow$	 & & & & & & & & & \\
		\hline \hline
		IUV Map & 125.2 & 125.9 & 128.3 & 132.3 & 133.6 & 136.8 & 144.0 & 144.3 & 191.5 \\ 
		Body Segment & 124.8 & 126.7 & 128.9 & 131.3 & 132.3 & 135.9 & 143.0 & 148.5 & 196.7 \\ 
		Image & 127.4 & 128.4 & 132.2 & 134.6 & 136.0 & 143.3 & 149.9 & 152.2 & 203.2 \\ 
		\hline
		Image \& IUV & 125.5 & 126.2 & 130.1 & 131.6 & 135.3 & 135.9 & 140.6 & 148.0 & 197.0 \\ 
		Image \& Body Segment & 125.8 & 126.1 & 129.5 & 131.4 & 133.7 & 136.5 & 143.3 & 148.0 & 196.4 \\ 
		\hline					
	\end{tabular}	
	\label{tab:3D_annotation}
\end{table*}

\begin{table*}[t] \centering  
	\caption{\textbf{Influence of constrained 3D annotations.}. The inputs of the models are all single images.}
	\setlength\tabcolsep{2pt}
	\begin{tabular}{c|c|c|c|c}
		\hline
		\multirow{2}{*}{Other Supervisions $\rightarrow$}  &   $100\%$ 3D \&   &  $20\%$ 3D \&  	&  Dense \&      &   Sparse 2D \\ 
		&	  Sparse 2D     &   Sparse 2D 	 	& Sparse 2D	     &     Only    \\
		\hline
		with Constained 3D              			&		127.4		&	137.7		    &	137.3		 &     	203.2	\\   
		\hline            
		w/o Constrained 3D							&		128.9		&	138.1			&	173.4     & 		230.9 \\
		\hline
	\end{tabular}	
	\label{tab:constrained}
\end{table*}

\end{document}